\date{}
\documentclass[12pt]{article}
\usepackage{graphicx}
\usepackage{amsmath}
\usepackage{amssymb}
\usepackage{subfigure}
\usepackage{multirow}
\usepackage{array}
\usepackage[margin=0.8in]{geometry}
\usepackage{algorithm}
\usepackage{algorithmic}
\usepackage{caption}
\graphicspath{{./Figs/},{./Figs/Gaussian_measurements/celebA/},{./Figs/scattering_fashion_results/},{./Figs/scattering_mnist_results/},{./Figs/Gaussian/fashion/},{./Figs/Gaussian/mnist/},{./Figs/Gaussian/celebA/},{./Figs/Gaussian/quantitative/},{./Figs/Gaussian/gaussian_bm3d/},{./Figs/Gaussian/cars/},{./Figs/Gaussian/svhn/},{./Figs/scattering_media/},{./Figs/scattering_media/fashion_results/},{./Figs/scattering_media/mnist_results/},{./Figs/coded_diffraction/celebA/},{./Figs/coded_diffraction/cars/},{./Figs/coded_diffraction/svhn/},{./Figs/noise/}}

\newcommand{\R}{\mathbb{R}}

\newcommand{\vct}[1]{\boldsymbol{#1}}

\newcommand{\set}[1]{\mathcal{#1}}

\newcommand{\vx}{\vct{x}}
\newcommand{\vy}{\vct{y}}
\newcommand{\vz}{\vct{z}}
\newcommand{\setN}{\set{N}}

\begin{document}

\title{Robust Compressive Phase Retrieval via Deep Generative Priors}
\author{Fahad Shamshad, Ali Ahmed \\ \small{Dept. of Electrical Engg., Information Technology University, Lahore, Pakistan.} \\ \tt{\small{$\lbrace$fahad.shamshad, ali.ahmed$\rbrace$@itu.edu.pk }} }

\maketitle

\begin{abstract}
This paper proposes a new framework to regularize the highly ill-posed and non-linear phase retrieval problem through deep generative priors using simple gradient descent algorithm. We experimentally show effectiveness of proposed algorithm for random Gaussian measurements (practically relevant in imaging through scattering media) and Fourier friendly measurements (relevant in optical set ups). We demonstrate that proposed approach achieves impressive results when compared with traditional hand engineered priors including sparsity and denoising frameworks for number of measurements and robustness against noise.  Finally, we show the effectiveness of the proposed approach on a real transmission matrix dataset in an actual application of multiple scattering media imaging. 
\end{abstract}


\section{Introduction}

{\let\thefootnote\relax\footnote{{Preprint. Work in progress.}}}
This paper considers recovering real valued signal $\boldsymbol{x} \in \R^{n}$ from its magnitude measurements of the form 
\begin{equation} \label{eq:pr}
{y}_i = \vert \langle \boldsymbol{a}_i, \boldsymbol{x} \rangle \vert + n_i, \quad \text{for} \;\; i = 1,2,...,m,
\end{equation}
where $\boldsymbol{y} \in \mathbb{R}^{m}$
is measurement vector, $\boldsymbol{A} = [\boldsymbol{a}_1, \boldsymbol{a}_2, ... , \boldsymbol{a}_m]^{T}$ is measurement matrix  and $\boldsymbol{n} \in \mathbb{R}^{m}$ denotes noise perturbation. This problem is known as \textit{phase retrieval} and is encountered frequently in applications including X-ray crystallography \cite{millane1990phase, harrison1993phase}, astronomy \cite{fienup1987phase}, optics \cite{shechtman2015phase}, tomography, microscopy, array imaging \cite{chai2010array}, acoustics \cite{balan2010signal}, quantum mechanics \cite{paul1994phase} and ptychography \cite{qian2014efficient}, where it is extremely difficult or infeasible to measure phase information of signal while recording magnitude measurements is much easier. In its full generality, the inverse problem \ref{eq:pr} is severely ill-posed due to its non-linear and non-convex nature.

Traditional approaches to overcome the ill posedness of phase retrieval generally falls into two categories. $\textit{First}$ approach is to introduce redundancy into measurement system, where we take more measurements than dimension of true signal $\boldsymbol{x}$, i.e., $m>n$ usually in the form of oversampled Fourier transform \cite{bendory2017fourier}, short-time Fourier transform \cite{bendory2018non}, random Gaussian measurements \cite{waldspurger2018phase}, coded diffraction patterns using random masks or structured illuminations \cite{candes2015phase, jaganathan2015phase}, wavelet transform \cite{waldspurger2017phase}, and Gabor frames \cite{bojarovska2016phase}. $\textit{Second}$ approach is to exploit some known knowledge about true signal $\boldsymbol{x}$ (prior information) such as sparsity \cite{jaganathan2017sparse, pauwels2018fienup, jaganathan2013sparse} or non-negativity \cite{beinert2017non, ahn1994relaxation}. Priors based approaches have recently attained much attention in phase retrieval community specially for the purpose of reducing number of measurements $m$ (compressive phase retrieval (CPR)) as acquisition of measurements is usually expensive and time consuming especially for large specimens at high resolutions \cite{ohlsson2012cprl,schniter2015compressive,metzler2016bm3d,pedarsani2017phasecode,zhang2017compressive}.

Most widely used prior for phase retrieval is sparsity as natural signals, especially images, are sparse or exhibit sparse representation in some known basis like Fourier or wavelets \cite{mallat2008wavelet}. Inspired from the theory of compressed sensing  \cite{donoho2006compressed, candes2008introduction}, many recent works have extended traditional phase retrieval algorithms (usually categorized into convex and non-convex algorithms) to incorporate sparse structure of underlying true signal for designing efficient recovery algorithms for compressive phase retrieval. Popular CPR algorithms include CPRL \cite{ohlsson2012cprl}, GESPAR \cite{shechtman2014gespar}, TSPR \cite{jaganathan2017sparse}, modified Fineup algorithm \cite{mukherjee2014fienup}, and sparse wirtinger flow \cite{yuan2017phase}. However, as shown in \cite{metzler2016bm3d} simple sparsity priors often fail to capture complicated structure that many natural signals exhibit resulting in unrealistic signals also fitting the sparse prior model assumption. Moreover, designing computationally efficient recovery algorithms for CPR is also very challenging. This led to the integration of more refined priors into phase retrieval problem , such as structured sparsity \cite{jagatap2017sample, jagatap2017fast}, dictionary models \cite{tillmann2016dolphin}, compression algorithms \cite{bakhshizadeh2017compressive}, and  total variation \cite{chang2016phase}. While these refined priors often provide superior performance compared to standard sparsity-based methods, they still suffer from the aforementioned limitations on modelling capability.

To capture complex natural structure of images recently plug-and-play and deep learning priors have been shown to produce state of the art results in many imaging linear inverse problems including denoising, super-resolution, inpainting, compressed sensing, etc. Following success of plug-and-play and deep learning priors recent works have extended them for solving phase retrieval problem. Plug-and-play methodologies for phase retrieval includes SPAR \cite{katkovnik2017phase}, BM3D-prgamp \cite{metzler2016bm3d} and plug-and-play ADMM \cite{venkatakrishnan2013plug, heide2016proximal}. Among them most popular is BM3D-prgamp algorithm that incorporates BM3D denoiser as prior in general message passing algorithm to solve CPR problem and have been shown to produce comparable results with far fewer noisy measurements when compared with sparsity prior.  Deep learning based approaches for solving phase retrieval are so far application specific that include holography \cite{rivenson2018phase}, Ptychography \cite{kappeler2017ptychnet}, etc. As these methods are application specific so they do not generalize to new phase retrieval set ups and require retraining of neural networks for different phase retrieval set ups and different noise levels.

\begin{figure*}[t]
\centering
\includegraphics[height = 5cm, width = 18cm]{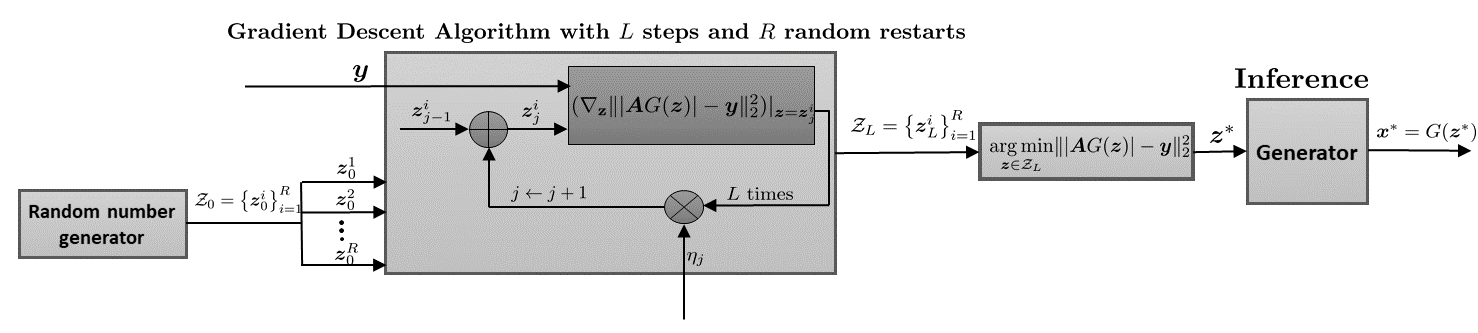}
\caption{\small{Overview of phase retrieval algorithm using deep generative prior. We use $R$ random restarts in our experiments. For each random restart gradient descent algorithm is run for $L$ steps to minimize the objective function  $ \Vert \vert \boldsymbol{A}G(\boldsymbol{z}) \vert - \boldsymbol{y} \Vert_2^{2}$ using gradient descent algorithm  where $\eta_j$ is step size at $j^{th}$ iteration. For each random restart, we save the latent vectors $\boldsymbol{z}$ of last iteration $L$ of gradient descent in set $\mathcal{Z}_L$. Among these $R$ latent vectors we choose $\boldsymbol{z}^{\ast}$ that give minimum error $ \Vert \vert \boldsymbol{A}G(\boldsymbol{z}) \vert - \boldsymbol{y} \Vert_2^{2}$. Optimal $\boldsymbol{z}$ denoted as $\boldsymbol{z}^{\ast}$ is given as input to generator in inference step to produce estimated image  $\boldsymbol{x}^*$.}}
	\label{fig:proposed_approach1}
\end{figure*}

Recently, neural networks based implicit generative models such as generative adverserial networks (GANs) \cite{goodfellow2014generative} have found success in modelling complex data distributions especially that of images.  GAN consists of two neural networks, generator ($G$) and discriminator ($D$). Generator tries to learn mapping from low dimensional latent space to points in space of high dimensional data (training data). On the other hand, $D$ tries to distinguish
real samples in the training dataset from fake samples synthesized by the $G$. Both $G$ and $D$ are trained simultaneously using backpropagation algorithm to the point where $G$ successfully fooled $D$ by generating fake data that is indistinguishable from real data. Due to their power of modelling natural images distributions, GANs have been extensively used to solve ill-pose linear inverse problems like compressed sensing \cite{bora2017compressed}, denoising \cite{mixon2018sunlayer} image superresolution \cite{ledig2017photo}, image inpainting \cite{yeh2017semantic}, blind deconvolution \cite{asim2018solving} etc. 

In this paper, we empirically show that given a noisy, phaseless measurements $\boldsymbol{y}$, and the assumption that the true image $\boldsymbol{x}$ belongs to the range of pre-trained generator, we can recover true image using simple gradient descent algorithm. Specifically, the algorithm searches for $\boldsymbol{x}$ in the range of pre-trained generator of images that explain the phaseless measurements $\boldsymbol{y}$. Since the range of the generative models can be traversed by a much lower dimensional latent representations compared to the ambient dimension of the images, it not only reduces the number of unknowns in the phase retrieval problem but also allows for an efficient implementation of gradients in this lower dimensional space using back propagation through the generators. Our numerical experiments manifest that, in general, the deep generative priors yield better results from far fewer measurements when compared with classical image priors.

In final stages of this work we came to know about work of \cite{paul2018pg} that solves the phase retrieval problem using generative priors with rigorous theoretical guarantees for random  Gaussian measurement matrix. We on the other hand provide rigorous experimental evaluation. In addition to Gaussian sensing matrix as in \cite{paul2018pg}, we provide extensive experimental evaluation with coded diffraction pattern measurements that are practically relevant in many optical settings. Experimental analysis for noise robustness of proposed approach and experiments with real transmission matrix dataset in an actual application of multiple scattering media imaging  \cite{metzler2017coherent} are also performed.

\subsection{Our Contributions}
Main contribution of our work is the combination of the powerful idea of deep generative model (GAN) with non-linear inverse problem of CPR for the first time. Specifically our work fits into recent trend of using advance priors for solving ill posed inverse problems. We show through extensive experiments that solving compressive phase retrieval problem using deep generative priors results in comparable performance to traditional prior based approaches with far fewer measurements. The resultant problem can be effectively solved using simple gradient descent scheme yielding promising results. It also turns out that using generative maps induces a very strong prior that is highly robust to noise. We show the effectiveness of proposed algorithm for coded diffraction pattern measurements (CDP) that are practically relevant in many optical setups. Finally, we demonstrate the effectiveness of proposed approach for imaging through scattering media on real measurement matrix dataset provided by \cite{metzler2017coherent}.

The rest of the paper is organized as follows. We formulate the problem with proposed approach in Section \ref{sec:Problem-Formulation}. Details of network architecture for GANs and description of datasets are provided in Section \ref{sec:numerical_experiments}. Section \ref{sec:gaussian} and Section \ref{sec:cdp} contains experimental results for random Gaussian and coded diffraction pattern measurement matrices, respectively. Section \ref{sec:noise_robust} shows performance of proposed approach for high additive noise. Section \ref{sec:scatter} provide details of experiments for multiple scattering media imaging on real transmission matrix dataset.

\section{Problem Formulation and Proposed Solution}\label{sec:Problem-Formulation}

We assume that image $\boldsymbol{x} \in \R^n$  in \eqref{eq:pr} is member of some structured class, denoted by $\mathcal{X}$, of images. Imaging applications of phase retrieval include coded diffraction imaging (CDI), astronomical imaging, X-ray imaging and optical imaging etc. Generative model represented by mapping $G: \mathbb{R}^k \rightarrow \mathbb{R}^n$ where $k \ll n$ is trained on representative sample set from class $\mathcal{X}$. Given low-dimensional input vector $\boldsymbol{z} \in \mathbb{R}^k$, the generator $G$ after training, generate new samples $G(\boldsymbol{z})$ similar to representative sample of the class $\mathcal{X}$. This generator is fixed after training (pre-trained generator). To recover the clean image, from the magnitude only measurements $\boldsymbol{y}$ in \eqref{eq:pr}, we propose minimizing the following objective function 
\begin{align}\label{eq:Optimization-Ambient}
{\boldsymbol{x}^*} :=  \underset{\substack{\boldsymbol{x} \in\text{Range}(G)}}{\text{argmin}} \ \|\boldsymbol{y} - \vert \boldsymbol{A} \boldsymbol{x} \vert \|^2, 
\end{align}
where Range($G$) is the set of all the images that can be generated by pre trained $G$. In other words, we want to find an image $\boldsymbol{x}$ that best explains the model  \eqref{eq:pr} and lies with in the range of generator. Ideally, the range of a generator comprises of only the samples drawn from the distribution of the images, i.e., $\boldsymbol{x} = G(\boldsymbol{z})$. Constraining the solution ${\boldsymbol{x}^*}$ to lie only in generator range, therefore, implicitly reduces the solution ambiguities, and forces the solution to be member of image class $\mathcal{X}$. 

The minimization program in \eqref{eq:Optimization-Ambient} can be equivalently formulated in the lower dimensional, latent representation space as follows
\begin{align}\label{eq:Optimization-latent}
\boldsymbol{z}^* = \underset{\boldsymbol{z} \in \R^k}{\text{argmin}}
\ 
\| \boldsymbol{y} - \vert \boldsymbol{A}G(\boldsymbol{z}) \vert \|^2,
\end{align}
For brevity we denote the objective by 
\begin{align}\label{eq:Optimization-latent}
\mathcal{L}(\boldsymbol{z}) =  
\| \boldsymbol{y} - \vert \boldsymbol{A}G(\boldsymbol{z}) \vert \|^2,
\end{align}
This optimization program can be thought of as tweaking the latent representation vector $\boldsymbol{z}$ (input to the $G$)  until this generator generate an image $\boldsymbol{x}$ that is consistent with \ref{eq:Optimization-latent}. 

The optimization program in \eqref{eq:Optimization-latent} is non-convex and non-linear owing to the modulus  operator, and non-linear deep generative model. We resort to gradient descent algorithm to find a local minima  $\boldsymbol{z}^*$. Importantly, the weights of the generator are always fixed as they enter into this algorithm as pre-trained models. The estimated image is acquired by a forward pass of the solution $\boldsymbol{z}^*$ through the generator $G$.  

Proposed algorithm is illustrated in Figure \ref{fig:proposed_approach1}. Note that due to its non-convex nature \ref{eq:Optimization-latent} can stuck in local minimas instead of reaching true global minima. Different initialization techniques have been proposed to find good initial guess of phase retrieval algorithm that will guarantee convergence. We circumvent this issue by running $L$ gradient descent steps from $R$ different random initialization of $\boldsymbol{z}$ denoted in Figure \ref{fig:proposed_approach1} by set $\mathcal{Z}_0$ where  $\mathcal{Z}_0 = \lbrace \boldsymbol{z}_0^i\rbrace_{i=1}^{R}$. Optimal $\boldsymbol{z}$, denoted as $\boldsymbol{z}^*$, is the one that gives minimum reconstruction error. The desired solution is $\boldsymbol{x}^* = G(\boldsymbol{z^*})$. 

\begin{algorithm} \label{alg:AltGradDescent}
	\caption{Phase retrieval via generative prior}
	\begin{algorithmic} 
		\STATE \textbf{Input:} $\vy$, $G$ and $\eta$ \\
		\textbf{Output:}  Estimates $\hat{\vx}$  \\
		\STATE \textbf{Initialize:}\\
		$\vz_0 :=  \setN(0,I_K)$ 
		\FOR{${t = 1,2,3,\ldots L}$}
		\STATE{$ \boldsymbol{z}_{t+1} \leftarrow \boldsymbol{z}_{t}$ - $\eta \nabla_{\vz_{t}} \mathcal{L}(\boldsymbol{z}_{t})$; }
		\ENDFOR \\
		${\vz}^{*}  \leftarrow G(\vz_L)$
   \end{algorithmic}
\end{algorithm}





\section{Numerical Simulations} \label{sec:numerical_experiments}
In this section, we evaluate performance of proposed algorithm, that we termed as PRGAN (Phase Retrieval using Generative Adversarial Network), with existing phase retrieval algorithms under different conditions\footnote{We have used Keras deep learning library with tensorflow backend in all our experiments. Code for reproducing subset of experiments will be released soon. Complete code for reproducing all experimental results will be released soon.}. In all our experiments  signal $\boldsymbol{x}$ is an image having resolution of $n$ pixels and our goal is to stably recover that from as few measurements $m$ as possible. All simulations are performed on core-i7 computer (3.40 GHz and 16GB RAM) equipped with Nvidia TITAN X GPU. 


 We assessed performance of PRGAN using peak signal to noise ratio (PSNR) and structural similarity index measure (SSIM).


\subsubsection{\textbf{Datasets description}}
We evaluate performance of PRGAN on five datasets, two grayscale and three color (RGB) datasets. Grayscale datasets include MNIST \cite{deng2012mnist} and Fashion MNIST \cite{xiao2017fashion} where as color datasets include CelebA \cite{liu2015deep}, SVHN \cite{netzer2011reading} and stanford cars \cite{krause2013collecting}. MNIST consists of $28 \times 28$ grayscale images of handwritten digits having 10 categories from $0 - 9$ with 60,000 training and 10,000 test examples. Fashion MNIST consists of $28 \times 28$ images with 10 categories of fashion products and 60,000 training and 10,000 test examples. 
RGB datasets include  CelebA contains more than $200,000$ RGB face images of size $218 \times 178 \times 3$ of different celebrities. We used alligned and cropped version of this dataset, where each image is of size $64 \times 64$. SVHN consists of $32 \times 32 \times 3$ real world images of house numbers obtained from google street view images with 73257 digits for training and 26032 digits for testing. We upsample SVHN to size $64 \times 64 \times 3$ and perform experiments on that dataset. The stanford cars dataset contains 16,185 images of different sizes. The data is split into 8,144 training images and 8,041 testing images and contain 196 different car models. We resize all images to size $64 \times 64 \times 3$ after segmenting cars regions using provided bounding boxes. Except stanford cars dataset, we use standard training dataset for training GAN. For cars dataset we use first 1000 images of standard test set as test images and train GAN on remaining images. 
 All experiments including comparison methods are performed by adding $1\%$\footnote{For an image scaled between 0 and 1, Gaussian noise of $1\%$ translate to Gaussian noise with standard deviation $\sigma = 0.01$ and mean $\mu=0$.} Gaussian noise to $\boldsymbol{y}$.

\begin{figure*}[t]
\centering

\subfigure{\rotatebox{90}{\hspace{0mm} \small Original} \hspace{0.05mm}\includegraphics[height = 2.65cm, width = 2.65cm]{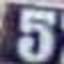}} 
 \subfigure{\includegraphics[height = 2.65cm, width = 2.65cm]{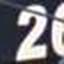}} \hspace{0.05mm}
  \subfigure{\includegraphics[height = 2.65cm, width = 2.65cm]{1-celebA.jpg}}
 \subfigure{\includegraphics[height = 2.65cm, width = 2.65cm]{2-celebA.jpg}} \hspace{0.05mm}
  \subfigure{\includegraphics[height = 2.65cm, width = 2.65cm]{test_car3.jpg}}
 \subfigure{\includegraphics[height = 2.65cm, width = 2.65cm]{test_car5.jpg}} \\[-0.7em]

\subfigure{\rotatebox{90}{\hspace{0mm} \small Range} \hspace{0.05mm}\includegraphics[height = 2.65cm, width = 2.65cm]{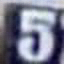}} 
\subfigure{\includegraphics[height = 2.65cm, width = 2.65cm]{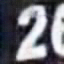}} \hspace{0.05mm}
\subfigure{\includegraphics[height = 2.65cm, width = 2.65cm]{1_closest_range.png}}
\subfigure{\includegraphics[height = 2.65cm, width = 2.65cm]{2_closest_range.png}} \hspace{0.05mm}
\subfigure{\includegraphics[height = 2.65cm, width = 2.65cm]{3range_cars.png}}
\subfigure{\includegraphics[height = 2.65cm, width = 2.65cm]{5range_cars.png}}  \\[-0.7em]

\subfigure{\rotatebox{90}{\hspace{1mm} \small BM3D} \hspace{0.05mm}\includegraphics[height = 2.65cm, width = 2.65cm]{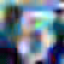}} 
\subfigure{\includegraphics[height = 2.65cm, width = 2.65cm]{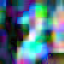}} \hspace{0.05mm}
\subfigure{\includegraphics[height = 2.65cm, width = 2.65cm]{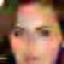}}
\subfigure{\includegraphics[height = 2.65cm, width = 2.65cm]{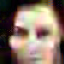}} \hspace{0.05mm}
\subfigure{\includegraphics[height = 2.65cm, width = 2.65cm]{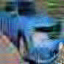}}
\subfigure{\includegraphics[height = 2.65cm, width = 2.65cm]{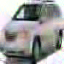}} \\[-0.7em]

\subfigure{\rotatebox{90}{\hspace{1mm} \small PRGAN} \hspace{0.05mm}\includegraphics[height = 2.65cm, width = 2.65cm]{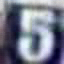}} 
\subfigure{\includegraphics[height = 2.65cm, width = 2.65cm]{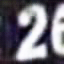}} \hspace{0.05mm}
\subfigure{\includegraphics[height = 2.65cm, width = 2.65cm]{500celebA1.png}}
\subfigure{\includegraphics[height = 2.65cm, width = 2.65cm]{500celebA2.png}} \hspace{0.05mm}
\subfigure{\includegraphics[height = 2.65cm, width = 2.65cm]{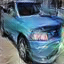}}
\subfigure{\includegraphics[height = 2.65cm, width = 2.65cm]{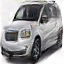}}

\caption{ \small{Results of our proposed approach PRGAN (Phase Retrieval using Generative Adversarial Network) on SVHN, CelebA and cars dataset for random complex Gaussian measurement matrix. For SVHN, CelebA, and cars dataset we show recovery results of PRGAN for 400, 500, and 1000 measurements respectively. We show original images (top row), closest images in range of generator (second row), reconstruction by BM3D-prgamp algorithm (third row) and reconstruction by PRGAN (last row).}}
\label{fig:celebA_gaussian}
\end{figure*}

\begin{figure*}[t]
\subfigure{\rotatebox{90}{\hspace{17mm} \small CelebA} \hspace{0.05mm}\includegraphics[height = 4.4cm, width = 17cm]{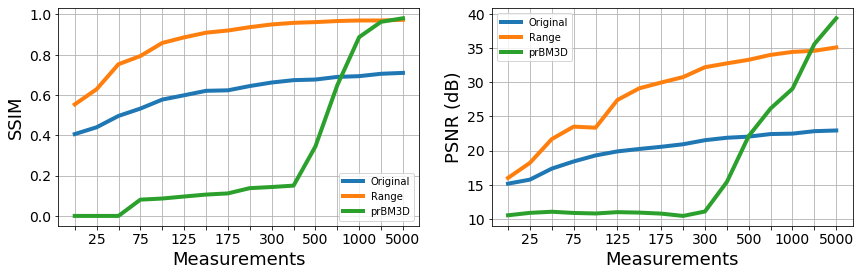}}
\subfigure{\rotatebox{90}{\hspace{17mm} \small SVHN} \hspace{0.05mm}\includegraphics[height = 4.4cm, width = 17cm]{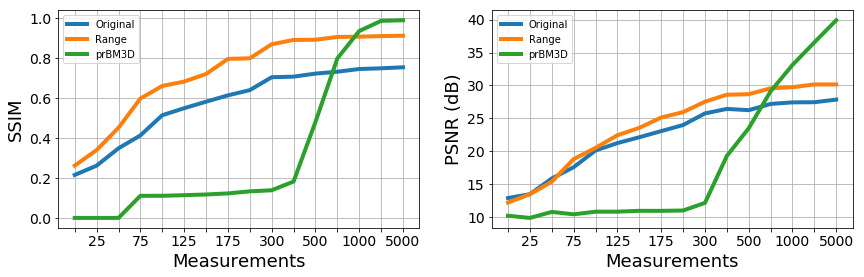}}
\subfigure{\rotatebox{90}{\hspace{18mm} \small Cars} \hspace{0.05mm}\includegraphics[height = 4.4cm, width = 17cm]{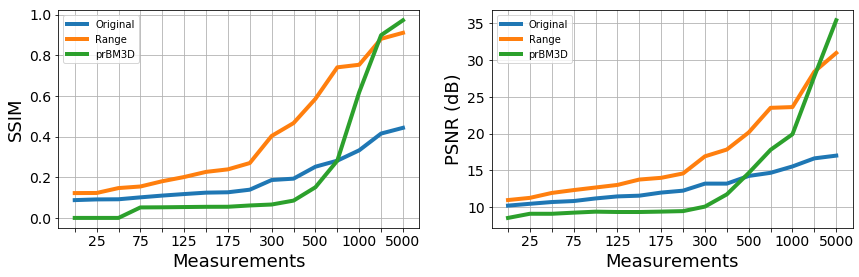}}
\caption{ \small{Quantitative results of proposed approach, PRGAN, in terms of PSNR, and SSIM with number of measurements for random complex Gaussian matrix. Results for CelebA, SVHN, and cars dataset for both original images and range images are shown in first, second, and third row, respectively. Compare to BM3D-prgamp, PRGAN is able to reconstruct images from far fewer measurements.}}
\label{fig:gaussian quantitative}
\end{figure*}

\subsubsection{\textbf{Generator architecture}}

\begin{table}[t]
\centering
\scalebox{0.9}{
\setlength{\arrayrulewidth}{.1mm}
\begin{tabular}{ ||c|c||}
\hline
\textbf{Generator} & \textbf{Discriminator} \\
\hline
 \textbf{Input:} Noise vector $\boldsymbol{z} \sim  \setN(0,I)$  & \textbf{Input:} Image $28^*28^*1$ \\[0.5ex] 
\hline 
MLP 1024, Batch norm, ReLU & Conv2D (filters 64, size 5, stride 2) \\
                           & Batch norm, ELU \\
\hline
MLP 6272, Batch norm, ReLU & Conv2D (filters 128, size 5, stride 2)\\
Reshape $7^*7^*128$        & Batch norm, ELU\\
\hline
Upsamp-2 & Maxpool (size 2), flatten \\
Conv2D (filters 64, size 5,stride 1) & MLP 256 \\
Batch norm, ReLU           &  Batch norm, Dropout 0.5, ELU\\
\hline 

Upsamp-2 &  \\
Conv2D (filters 1, size 5, stride 1) & MLP 1, Sigmoid \\
Batch norm, Tanh & \\

\hline
\end{tabular}}
\caption{ \small{Generator and discriminator architectures used for training of MNIST and Fashion MNIST dataset.}}
\label{table:1}
\end{table}

For grayscale datasets, architectures for generator and discriminator are given in Table \ref{table:1}. Size of low-dimensional vector $\boldsymbol{z}$ is $40$ and sampled from random uniform distribution. Adam optimizer has been used for training with learning rate $0.0002$, $\beta_1 = 0.5$, batch size 32 and number of epochs 50.
 For RGB datasets we use deep convolutional generative adversarial network (DCGAN) model of \cite{radford2015unsupervised}. Size of low dimensional feature representation $\boldsymbol{z}$ is set to $100$ and sampled from random normal distribution. DCGAN model is trained by updating generator $\boldsymbol{G}$  twice and discriminator $\boldsymbol{D}$ once in each cycle to avoid fast convergence of $\boldsymbol{D}$.  Each update during training used the Adam optimizer \cite{kingma2014adam} with batch size 64, $\beta_1 = 0.5$, and learning rate $0.0002$. For all experiments we use $\lambda = 0.001$ and $\gamma = 0.001$. We use $10$ random restarts and choose estimate with minimum loss as our final estimate. Gradient descent is used as optimizer with total iterations of $10,000$ and a fixed step size of $0.001$.

\section{Gaussian measurements} \label{sec:gaussian}
 In this section, we evaluate the performance of PRGAN algorithm with random complex Gaussian measurements. Specifically, we show how the performance of our algorithm depends on the number of measurements. Although random measurements are far from structured measurements that one encounter in X-ray imaging and related applications \cite{candes2015phase} however recently these random measurements have found applications in imaging through multiple scattering media as will be described in Section \ref{sec:scatter} \cite{liutkus2014imaging,rajaei2016intensity,metzler2017coherent}.
\begin{figure*}[t]
\centering
\subfigure{\rotatebox{90}{\hspace{0mm} \small Original} \hspace{0.05mm}\includegraphics[height = 2.65cm, width = 2.65cm]{0-celebA.jpg}} 
 \subfigure{\includegraphics[height = 2.65cm, width = 2.65cm]{1-celebA.jpg}}
\subfigure{\includegraphics[height = 2.65cm, width = 2.65cm]{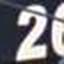}} 
 \subfigure{\includegraphics[height = 2.65cm, width = 2.65cm]{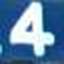}}
 \subfigure{\includegraphics[height = 2.65cm, width = 2.65cm]{test_car0.jpg}} 
 \subfigure{\includegraphics[height = 2.65cm, width = 2.65cm]{test_car3.jpg}} \\[-0.7em]
\subfigure{\rotatebox{90}{\hspace{0mm} \small Range} \hspace{0.05mm}\includegraphics[height = 2.65cm, width = 2.65cm]{0_closest_range.png}} 
\subfigure{\includegraphics[height = 2.65cm, width = 2.65cm]{1_closest_range.png}}
\subfigure{\includegraphics[height = 2.65cm, width = 2.65cm]{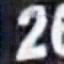}} 
\subfigure{\includegraphics[height = 2.65cm, width = 2.65cm]{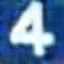}}
\subfigure{\includegraphics[height = 2.65cm, width = 2.65cm]{0range_cars.png}} 
\subfigure{\includegraphics[height = 2.65cm, width = 2.65cm]{3range_cars.png}} \\[-0.7em]
\subfigure{\rotatebox{90}{\hspace{1mm} \small BM3D} \hspace{0.05mm}\includegraphics[height = 2.65cm, width = 2.65cm]{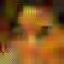}} 
\subfigure{\includegraphics[height = 2.65cm, width = 2.65cm]{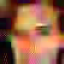}}
\subfigure{\includegraphics[height = 2.65cm, width = 2.65cm]{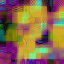}} 
\subfigure{\includegraphics[height = 2.65cm, width = 2.65cm]{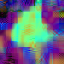}}
\subfigure{\includegraphics[height = 2.65cm, width = 2.65cm]{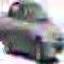}} 
\subfigure{\includegraphics[height = 2.65cm, width = 2.65cm]{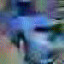}}

\subfigure{\rotatebox{90}{\hspace{1mm} \small PRGAN} \hspace{0.05mm}\includegraphics[height = 2.65cm, width = 2.65cm]{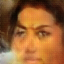}} 
\subfigure{\includegraphics[height = 2.65cm, width = 2.65cm]{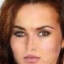}}
\subfigure{\includegraphics[height = 2.65cm, width = 2.65cm]{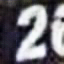}} 
\subfigure{\includegraphics[height = 2.65cm, width = 2.65cm]{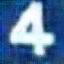}}
\subfigure{\includegraphics[height = 2.65cm, width = 2.65cm]{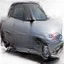}} 
\subfigure{\includegraphics[height = 2.65cm, width = 2.65cm]{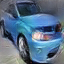}}
\caption{ \small{Results of our proposed approach PRGAN (Phase Retrieval using Generative Adversarial Network) on CelebA, SVHN and cars dataset for coded diffraction pattern measurements. For CelebA, SVHN, and cars dataset, we show recovery results of PRGAN for 400, 500, and 750 measurements, respectively. We show original images (top row), closest images in range of generator (second row), reconstruction by BM3D-prgan algorithm (third row) and reconstruction by PRGAN (last row). }}
\label{fig:CelebA_cdp}
\end{figure*}


For random measurements we have compared performance of PRGAN with compressive phase retrieval algorithm BM3D-prgamp \cite{metzler2016bm3d} 
that have shown state of the art performance when compared with existing prior based compressive phase retrieval algorithms including prGAMP \cite{schniter2015compressive}.

For BM3D-prgamp we use 10 random restarts, each with 6000 iterations of gradient descent, and choose estimate that give minimum reconstruction error. Other parameters are set to their default values.  For RGB datasets we use CBM3D denoiser \cite{dabov2007color}, variant of BM3D,  that has been proposed specifically for RGB images.

In Figure \ref{fig:celebA_gaussian}, we show reconstruction results by proposed PRGAN algorithm and BM3D-prgamp method on SVHN, CelebA, and cars dataset. We observe visually that our results are much better when compared with BM3D-prgamp results that are blurry at fewer measurements. However as our algorithm is limited by range of generator so by further increasing measurements results in saturation of performance as shown in Figure \ref{fig:gaussian quantitative}. As BM3D-prgamp has no such limitation so increasing measurements result in increase in its performance that eventually surpass PRGAN performance as shown quantitatively in Figure \ref{fig:gaussian quantitative}.

\section{Coded Diffraction Pattern measurements} \label{sec:cdp}

\begin{figure}[H]
	\centering
	\includegraphics[height = 5cm, width = 15cm]{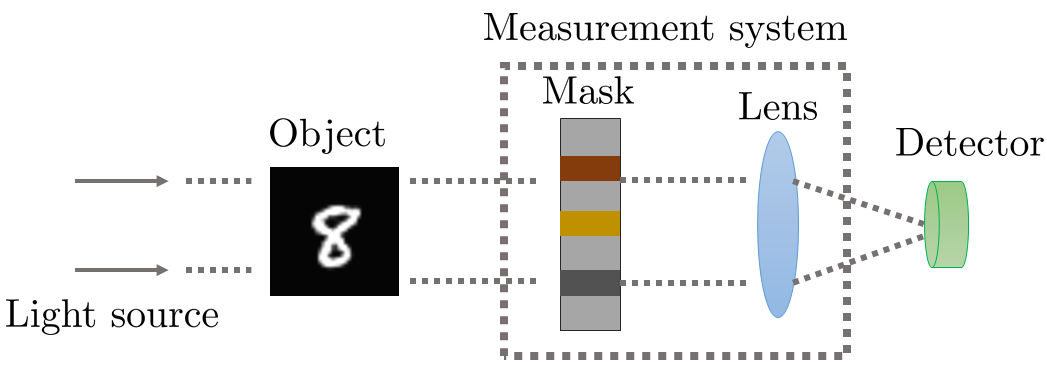}
	\caption{\small{Typical setup of optical imaging. Light through object of interest is passed through mask or coded diffraction pattern, and then through lens. }}
	\label{fig:cdp_setup}
\end{figure}

In many imaging applications such as optical imaging, measurement matrix cannot be arbitrary. In contrast to Gaussian measurements, in these applications measurement matrix is far more structured. A typical setup of optical imaging is shown in Figure \ref{fig:cdp_setup}. Light through object $\boldsymbol{x}$ is passed through mask followed by lens. Mathematically,  mask modulates each entry of $\boldsymbol{x}$ while lens take the Fourier transform of the modulated $\boldsymbol{x}$. So our measurement matrix $\boldsymbol{A}$ is 
\begin{equation}
\boldsymbol{A} = \boldsymbol{FD},
\end{equation} 
where $\boldsymbol{F}$ is 2D DFT matrix indicating operation performed by lens, $\boldsymbol{D}$ is diagonal mask matrix of size $N \times N$ having diagonal entries drawn uniformly from the unit circle in the complex plane of size. 
Measurements for total of $M$ masks  will be
 
\begin{equation}
\boldsymbol{A} = 
 \begin{bmatrix}
  \boldsymbol{FD}_1  \\
  \boldsymbol{FD}_2  \\
  \vdots \\
  \boldsymbol{FD}_M  
\end{bmatrix},
\end{equation}
For compressive phase retrieval we can write 
\begin{equation}
\boldsymbol{A} = 
 \begin{bmatrix}
  \boldsymbol{J}_1 \boldsymbol{FD}_1  \\
  \boldsymbol{J}_2 \boldsymbol{FD}_2  \\
  \vdots \\
  \boldsymbol{J}_M \boldsymbol{FD}_M 
\end{bmatrix},
\end{equation}
where $\boldsymbol{J}_i$ are selection matrices of size $m \times n$ and consist of $m$ randomly selected rows of $n \times n$ identity matrix.


As in Gaussian measurements case, for BM3D-prgamp we use 10 random restarts, each with 6000 iterations of gradient descent, and choose estimate that give minimum reconstruction error as solution.

We have compared performance of PRGAN in terms of number of measurements with BM3D-prgamp. Qualitative results for coded diffraction pattern measurements are shown in Figure \ref{fig:CelebA_cdp}. For CelebA, SVHN, and cars dataset we show recovery results of PRGAN for 400, 500, and 750 measurements, respectively from single diagonal mask drawn uniformly from the unit circle in the complex plane. Compared to BM3D-prgamp, PRGAN is able to reconstruct images from far fewer measurements. 
\section{Noise Robustness} \label{sec:noise_robust}
\begin{figure}[H]
\centering
\subfigure{\rotatebox{90}{\hspace{8mm} \small Original} \hspace{0.05mm}\includegraphics[height = 3.5cm, width = 3.5cm]{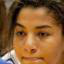}} 
 \subfigure{\includegraphics[height = 3.5cm, width = 3.5cm]{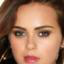}}
 \subfigure{\includegraphics[height = 3.5cm, width = 3.5cm]{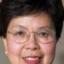}}
 \subfigure{\includegraphics[height = 3.5cm, width = 3.5cm]{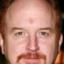}}  \\[-0.7em]
\subfigure{\rotatebox{90}{\hspace{8mm} \small Range} \hspace{0.05mm}\includegraphics[height = 3.5cm, width = 3.5cm]{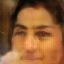}} 
\subfigure{\includegraphics[height = 3.5cm, width = 3.5cm]{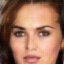}}
\subfigure{\includegraphics[height = 3.5cm, width = 3.5cm]{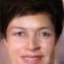}}
\subfigure{\includegraphics[height = 3.5cm, width = 3.5cm]{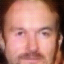}} \\[-0.7em]
\subfigure{\rotatebox{90}{\hspace{12mm} \small 5$\%$} \hspace{0.05mm}\includegraphics[height = 3.5cm, width = 3.5cm]{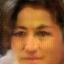}} 
\subfigure{\includegraphics[height = 3.5cm, width = 3.5cm]{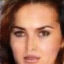}}
\subfigure{\includegraphics[height = 3.5cm, width = 3.5cm]{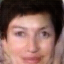}}
\subfigure{\includegraphics[height = 3.5cm, width = 3.5cm]{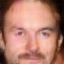}} \\[-0.7em]

\subfigure{\rotatebox{90}{\hspace{12mm} \small 25$\%$} \hspace{0.05mm}\includegraphics[height = 3.5cm, width = 3.5cm]{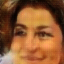}} 
\subfigure{\includegraphics[height = 3.5cm, width = 3.5cm]{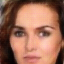}}
\subfigure{\includegraphics[height = 3.5cm, width = 3.5cm]{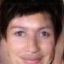}}
\subfigure{\includegraphics[height = 3.5cm, width = 3.5cm]{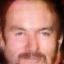}} \\[-0.7em]
\subfigure{\rotatebox{90}{\hspace{12mm} \small 50$\%$} \hspace{0.05mm}\includegraphics[height = 3.5cm, width = 3.5cm]{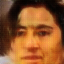}} 
\subfigure{\includegraphics[height = 3.5cm, width = 3.5cm]{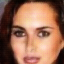}}
\subfigure{\includegraphics[height = 3.5cm, width = 3.5cm]{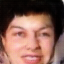}}
\subfigure{\includegraphics[height = 3.5cm, width = 3.5cm]{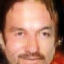}} \\[-0.7em]
\caption{ \small{Effectiveness of PRGAN algorithm against high additive noise. Results for CelebA dataset are shown with original images (top row), range images (second row), PRGAN recovery results with 5$\%$ (third row), 25$\%$ (fourth row), and 50$\%$ (last row). }}
\label{fig:CelebA_noise}
\end{figure}

\begin{figure}[h]
\centering
\subfigure{\includegraphics[height = 7.5cm, width = 13.5cm]{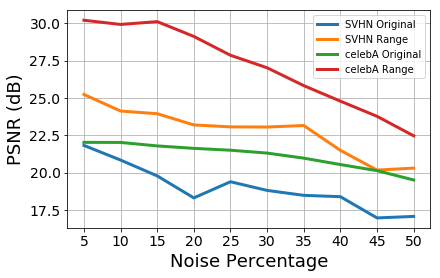}} \\[-0.4em]
\subfigure{\includegraphics[height = 7.5cm, width = 13.5cm]{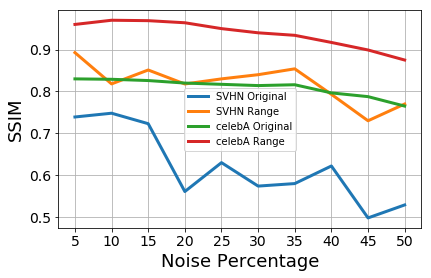}}  
\caption{ \small{Effectiveness of proposed approach PRGAN against additive noise for Gaussian measurement matrix. We show results in terms of PSNR, and SSIM for CelebA and SVHN datasets. }}
\label{fig:PRGAN_NOISE}
\end{figure}

 In this section, we evaluate the robustness of proposed PRGAN algorithm to additive Gaussian noise. The algorithm is robust to strong noise level as found in many imaging applications. Figure \ref{fig:CelebA_noise} and Figure \ref{fig:PRGAN_NOISE} demonstrate the effectiveness of PRGAN approach against additive noise even at high noise level of $50 \%$. Note that PRGAN algorithm is blind to noise level and type i.e, it does not require any information about noise level and noise type.

\section{Imaging through Multiple Scattering Media} \label{sec:scatter}
\begin{figure}[h]
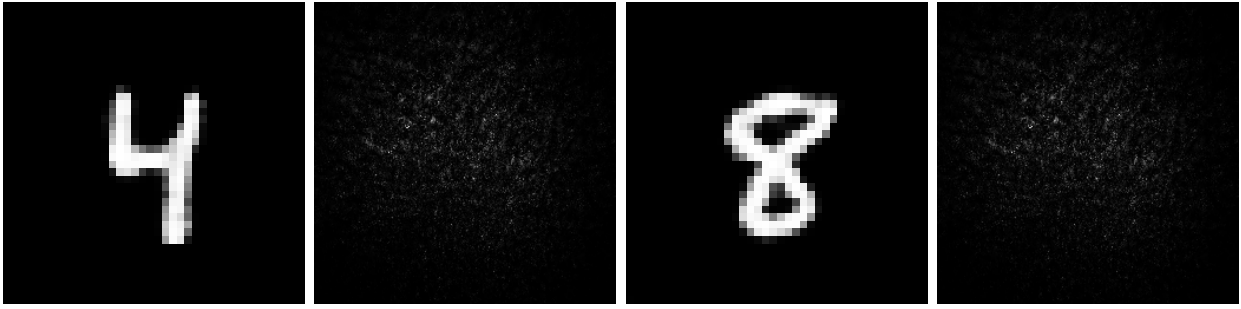

\centering

\subfigure{\includegraphics[height = 4cm, width = 4cm]{4.png}}
\subfigure{\includegraphics[height = 4cm, width = 4cm]{4_power_scatter.png}} 
\subfigure{\includegraphics[height = 4cm, width = 4cm]{8.png}}
\subfigure{\includegraphics[height = 4cm, width = 4cm]{4_power_scatter.png}} \\[-0.4em]
\caption{ \small{Digits along with their scattering pattern after multiplication with transmission matrix of turbulent media. The speckle pattern bears no resemblance with $\boldsymbol{x}$ thus making this problem highly ill-posed.}}
\label{fig:scatterer}
\end{figure}

Imaging through scattering or random media such as  glass diffuser or disordered nanoparticles is considered a challenging problem in computational optics due to rapid attenuation of light that prevent use of conventional imaging techniques. Key to solve this problem is to find the transmission matrix (TM) that characterizes the input-output relationship of light wavefront as it passes through scattering media. As the scattering process is linear so we can write
\begin{equation} \label{eq:scattering}
\boldsymbol{y} = \boldsymbol{Ax}+\boldsymbol{n},
\end{equation}
where $\boldsymbol{x}$ is coherent incident light that after passing through scattering medium with TM $\boldsymbol{A}$ produces speckle pattern $\boldsymbol{y}$ on far side of scatterer and $\boldsymbol{n}$ is  noise perturbation. The speckle pattern $\boldsymbol{y}$ bears no resemblance with $\boldsymbol{x}$ as shown in Figure \ref{fig:scatterer}.

\begin{figure}[h]
	\centering
	\includegraphics[height = 7.8cm, width = 14cm]{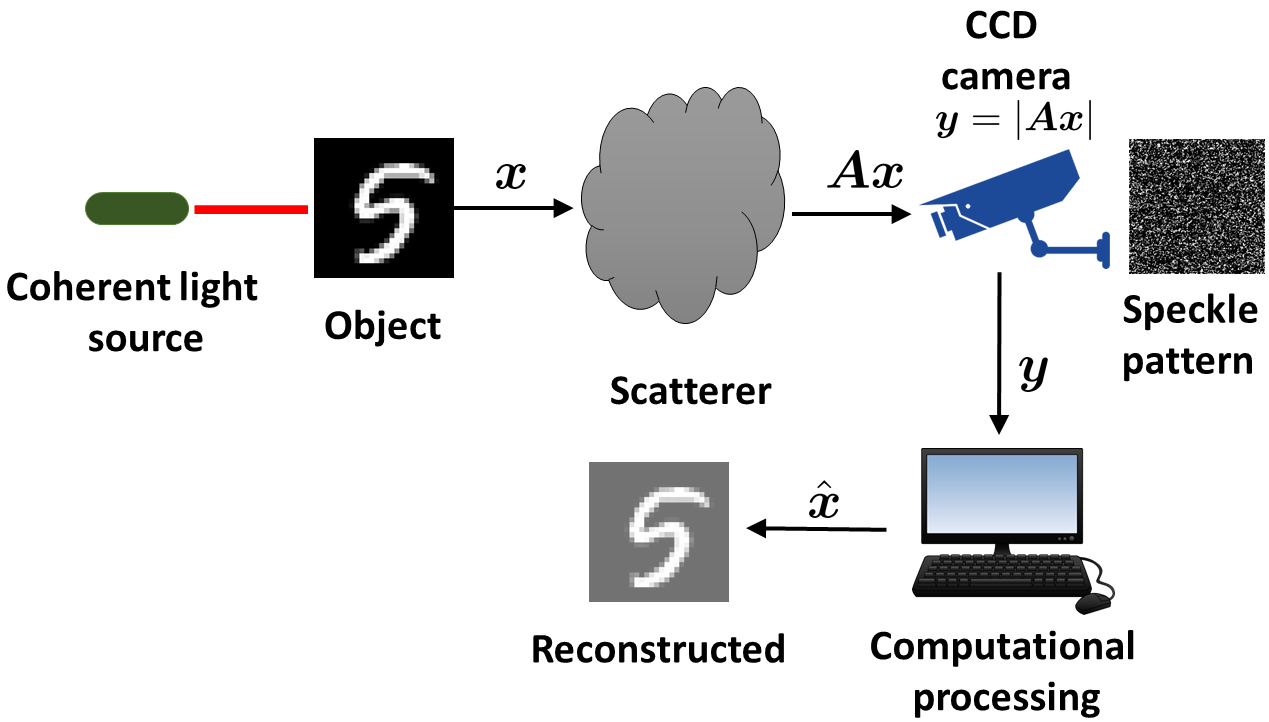}
	\caption{\small{Overview of imaging through scatterer. Object (here digit 5) is illuminated with coherent light source to produce signal $\boldsymbol{x}$ that passes through scatterer with transmission matrix $\boldsymbol{A}$ (glass diffuser, thin paint wall or biological tissue) to produce speckle pattern $\boldsymbol{y} = \boldsymbol{Ax}$ on far side of scatterer. Conventional cameras only capture intensity of light and thus with known TM, estimate $\boldsymbol{x}^*$ can be reconstructed with  phase retrieval algorithms.}}
	\label{fig:scattering_media}
\end{figure}

As typical cameras capture only intensity of light \cite{shechtman2015phase} so \ref{eq:scattering} can be written as
\begin{equation} \label{eq:scatterphase}
\boldsymbol{y}^2 = \vert \boldsymbol{Ax} + \boldsymbol{n}\vert^2.
\end{equation}
We take out square and will deal with $\boldsymbol{y} = \vert \boldsymbol{Ax} + \boldsymbol{n}\vert$. General setup for imaging through scattering media is shown in Figure \ref{fig:scattering_media}. Note that if we know the TM of scattering medium then problem \ref{eq:scatterphase} reduces to traditional phase retrieval problem and we can solve it using PRGAN algorithm.


Recently double phase retrieval algorithms have been developed for imaging through scattering media \cite{dremeau2015reference,rajaei2016intensity}. Compared to other methods of imaging through scattering media like time of flight \cite{heide2013low,tadano2015depth}, multiscale light propagation \cite{tian20153d}, temporallay modulated phase \cite{cui2011parallel}, strong memory effect \cite{katz2014non} and holographic interferometry \cite{popoff2010measuring} - this algorithm is cheap and easy to setup \cite{metzler2017coherent}.

Double phase retrieval approach gets its name from fact that phase retrieval algorithm is applied twice, once for calibration of TM of scattering media on training images and once for reconstruction of test images using estimated TM from first step through phase retrieval algorithm. Our work focus on later step. We will use real dataset for TM, provided by \cite{metzler2017coherent}, that use double phase retrieval approach for imaging through scattering media. Specifically they proposed fast phase retrieval algorithm prVAMP (based on previous VAMP-GLM \cite{schniter2016vector} algorithm) that runs hundreds of time faster than competing algorithms.
However their proposed algorithm does not incorporate any prior knowledge about target object and thus require $m \geq 4n -4 $ measurements \cite{bodmann2015stable} for reconstruction of $\boldsymbol{x}$ . We show experimentally that our proposed PRGAN algorithm is able to reconstruct target object from far fewer TM rows (selected randomly from original transmission matrix data) when compared with prVAMP approach using power of deep generative priors.
\subsection{TM Dataset:} TM data provided by \cite{metzler2017coherent} has dimensions of $256^2 \times 40^2$ where each row of TM is estimated using prVAMP algorithm. 
TM data  also contains the corresponding normalized residual vector for each row of TM with values ranging from $0.1 - 1$, that describes the accuracy of each estimated row of the provided TM. For experiments we only consider rows with residual values less than 0.4. As observed in \cite{metzler2017coherent} about 98$\%$ rows of TM have residual smaller than 0.4. 
\subsection{Experiments:} 
\begin{figure}[t]
	\centering
	\includegraphics[height = 6.5cm, width = 10cm]{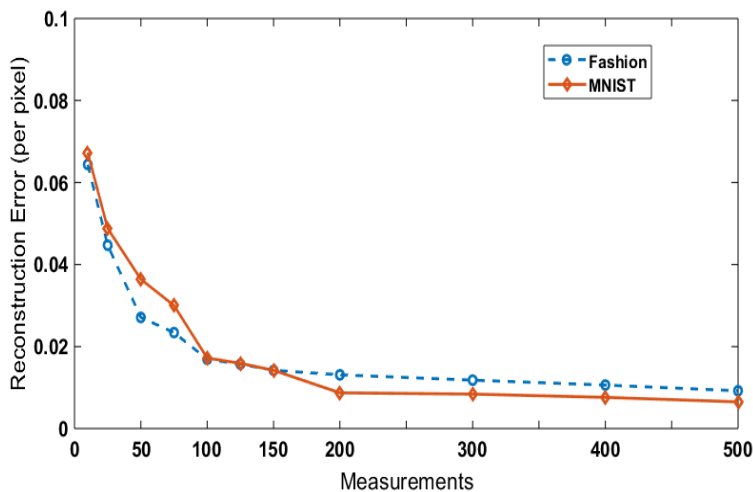}
	\caption{\small{Per pixel error for MNIST and Fashion MNIST dataset using PRGAN algorithm with transmission matrix of scatering media provided by \cite{metzler2017coherent} as measurement matrix with 300 measurements. Reconstruction error for 40 images from test set of MNIST and Fashion MNIST has been shown with number of measurements. }}
	\label{fig:perpixel_error}
\end{figure}

TM has been estimated for calibration patterns having dimension $40 \times 40$ so we zero pad training dataset of MNIST and Fashion MNIST to make it $40 \times 40$ and train GAN on this zero padded dataset. We use the same architecture of GAN as of Table \ref{table:1}. We randomly select 300 rows of TM having error residual less than 0.4 and use it as our measurement matrix. We randomly select 40 test images each from test set of MNIST and Fashion MNIST dataset for our experiments. As shown for Gaussian and coded diffraction pattern measurements the major source of error is due to true image not lie near or in the range of generator. We eliminate this error for given test image of MNIST or Fashion MNIST by finding corresponding closest image in $\ell_2$ distance that lie in the range of generator. PRGAN algorithm is then used for reconstruction for these range images. For 300 measurements results are shown in Figure \ref{fig:scattering_results}. Quantitative results in terms of per pixel error are shown in Figure \ref{fig:perpixel_error}. We can see that PRGAN can reconstruct almost perfect images from few TM measurements. 

\begin{figure}[H]
\centering

\subfigure{\rotatebox{90}{\hspace{5mm} \small Original} \hspace{0.05mm}\includegraphics[height = 2.7cm, width = 2.7cm]{0_test_sampled.png}} 
\subfigure{\includegraphics[height = 2.7cm, width = 2.7cm]{1_test_sampled.png}}
\subfigure{\includegraphics[height = 2.7cm, width = 2.7cm]{2_test_sampled.png}} 
\subfigure{\includegraphics[height = 2.7cm, width = 2.7cm]{3_test_sampled.png}} 
\subfigure{\includegraphics[height = 2.7cm, width = 2.7cm]{4_test_sampled.png}}
\subfigure{\includegraphics[height = 2.7cm, width = 2.7cm]{5_test_sampled.png}} \\[-0.7em]

\subfigure{\rotatebox{90}{\hspace{5mm} \small PRGAN} \hspace{0.05mm}\includegraphics[height = 2.7cm, width = 2.7cm]{120_meas0.png}} 
\subfigure{\includegraphics[height = 2.7cm, width = 2.7cm]{120_meas1.png}}
\subfigure{\includegraphics[height = 2.7cm, width = 2.7cm]{120_meas2.png}} 
\subfigure{\includegraphics[height = 2.7cm, width = 2.7cm]{120_meas3.png}} 
\subfigure{\includegraphics[height = 2.7cm, width = 2.7cm]{120_meas4.png}}
 \subfigure{\includegraphics[height = 2.7cm, width = 2.7cm]{120_meas5.png}}  \\[-0.7em]

\subfigure{\rotatebox{90}{\hspace{5mm} \small Original} \hspace{0.05mm}\includegraphics[height = 2.7cm, width = 2.7cm]{0_test_sampledm.png}} 
\subfigure{\includegraphics[height = 2.7cm, width = 2.7cm]{1_test_sampledm.png}}
\subfigure{\includegraphics[height = 2.7cm, width = 2.7cm]{2_test_sampledm.png}} 
\subfigure{\includegraphics[height = 2.7cm, width = 2.7cm]{3_test_sampledm.png}} 
\subfigure{\includegraphics[height = 2.7cm, width = 2.7cm]{4_test_sampledm.png}}
\subfigure{\includegraphics[height = 2.7cm, width = 2.7cm]{5_test_sampledm.png}} \\[-0.7em]

\subfigure{\rotatebox{90}{\hspace{5mm} \small PRGAN} \hspace{0.05mm}\includegraphics[height = 2.7cm, width = 2.7cm]{120_meas0m.png}} 
\subfigure{\includegraphics[height = 2.7cm, width = 2.7cm]{120_meas1m.png}}
\subfigure{\includegraphics[height = 2.7cm, width = 2.7cm]{120_meas2m.png}} 
\subfigure{\includegraphics[height = 2.7cm, width = 2.7cm]{120_meas3m.png}} 
\subfigure{\includegraphics[height = 2.7cm, width = 2.7cm]{120_meas4m.png}}
 \subfigure{\includegraphics[height = 2.7cm, width = 2.7cm]{120_meas5m.png}} 
\caption{ \small{Reconstruction results on MNIST and Fashion MNIST test images with transmission matrix provided by \cite{metzler2017coherent} using PRGAN algorithm with 300 measurements. Reconstructed images are shown in second and fourth row corresponding to original images in first and third row. We use 10 random restarts and show results with minimum reconstruction error as final.}}
\label{fig:scattering_results}
\end{figure}

\bibliographystyle{IEEEtran}
\bibliography{final}

\end{document}